# FloorplanMAE

*A self-supervised framework for complete floorplan generation from partial inputs*

Jun Yin[1], Jing Zhong[2], Pengyu Zeng[3], Peilin Li[4], Miao Zhang[5], Ran Luo[6], Shuai Lu[7*]
[1,2,3,5,7] *Tsinghua University*
[7]*shuai.lu@sz.tsinghua.edu.cn*
[*]*corresponding author*

**Abstract.** In the architectural design process, floorplan design is often a dynamic and iterative process. Architects progressively draw various parts of the floorplan according to their ideas and requirements, continuously adjusting and refining throughout the design process. Therefore, the ability to predict a complete floorplan from a partial one holds significant value in the design process. Such prediction can help architects quickly generate preliminary designs, improve design efficiency, and reduce the workload associated with repeated modifications. To address this need, we propose FloorplanMAE, a self-supervised learning framework for restoring incomplete floor plans into complete ones. First, we developed a floor plan reconstruction dataset, FloorplanNet, specifically trained on architectural floor plans. Secondly, we propose a floor plan reconstruction method based on Masked Autoencoders (MAE), which reconstructs missing parts by masking sections of the floor plan and training a lightweight Vision Transformer (ViT). We evaluated the reconstruction accuracy of FloorplanMAE and compared it with state-of-the-art benchmarks. Additionally, we validated the model using real sketches from the early stages of architectural design. Experimental results show that the FloorplanMAE model can generate high-quality complete floor plans from incomplete partial plans. This framework provides a scalable solution for floor plan generation, with broad application prospects.

## 1. Introduction

The process of drawing architectural floor plans is complex and challenging, requiring architects to gradually refine and perfect the design to meet specific project needs and creative visions. In particular, residential buildings, as a primary architectural type, often rely on highly trained architects for floor plan drawing and modifications, consuming substantial time and human resources. During this process, the automation of architectural design through predicting an entire floor plan from partial data becomes crucial. Not only can this assist architects in quickly generating preliminary design concepts, but it can also significantly improve design efficiency, reduce modification and rework time, and swiftly construct design concepts, thereby saving time and costs (Li et al., 2019).

While computer vision has developed rapidly (Zhang et al., 2024, 2025; Yin et al., 2024), modern models require large amounts of data and typically demand extensive labeled images (Zou et al., 2021). Natural images often contain a lot of redundant information, making it possible to infer missing parts from neighboring image blocks without the model needing to fully understand the objects and scene relationships (Huang et al., 2025). However, architectural floor plans are more structured, abstract, and diverse compared to natural images, meaning that the model must capture geometric structures and understand design logic and functional zoning , which poses additional challenges for generating architectural plans. To address these challenges, we first created a new, architectural floor plan reconstruction dataset—FloorplanNet. This dataset is specifically designed for the complexity and diversity of architectural plans, containing a variety of residential building layouts and floor plans, providing rich resources for training and evaluating models.

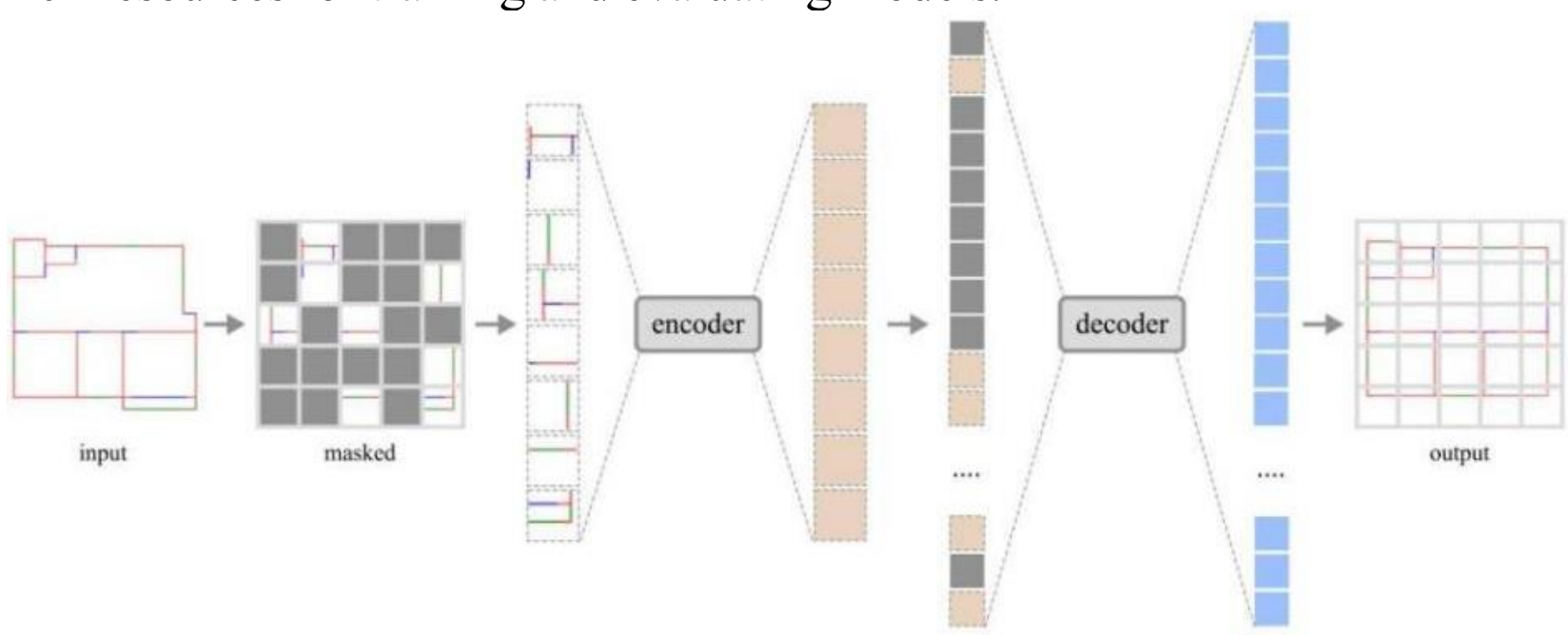

*Figure 1. FloorplanMAE Architecture Diagram.*

Next, we propose a floor plan reconstruction method based on mask autoencoder, as Figure 1 shows. This method masks parts of the input floor plan, then trains a lightweight Vision Transformer (ViT) to reconstruct the masked sections. In this way, the model learns to infer the complete floor plan structure from limited information. Moreover, this method employs a self-supervised learning approach, allowing the model’s understanding of images to go beyond simple statistical results. To validate the effectiveness of FloorplanMAE, we conducted a series of experiments. We evaluated the model's performance in reconstruction accuracy and compared it with current state-of-the-art benchmarks. Experimental results show that FloorplanMAE achieves excellent results in the reconstruction task, accurately restoring complete architectural floor plans from partial ones.

## 2. Related Work

### 2.1. APPLICATION OF SELF-SUPERVISED LEARNING

Self-supervised learning strategies have become a highly influential pretraining method in the fields of natural language processing (NLP) and computer vision (CV) . The core mechanism of this approach involves masking certain parts of the input data and training the model to predict the missing content as a means of learning the inherent structure of the data. In NLP, models like BERT (Devlin et al., 2018) and GPT (Radford, et al., 2018) use Masked Language Modeling (MLM) techniques, which involve randomly masking fragments of input text and training the model to predict these masked parts, successfully capturing deep semantic features of language. This pretraining paradigm not only enhances the model's generalized understanding of language. In the field of computer vision, self-supervised learning has similarly shown great potential. BEiT (Bao, et al., 2021) implements self-supervised pretraining by predicting masked discrete tokens within an image, similar to the MLM method in NLP. Although BEiT is inefficient in handling high-resolution images and requires a large amount of labeled data for fine-tuning, it provides a new perspective on image processing. These methods still face challenges when training on large-scale datasets, and designing suitable pretraining tasks requires domain-specific knowledge.

## 2.2. DEVELOPMENT OF AUTOENCODERS

Autoencoders (AEs) are a class of neural networks designed for learning compressed representations of data. They map input data to a lower-dimensional latent space through an encoder and use a decoder to reconstruct the original input. Denoising Autoencoders (DAEs) (Vincent, et al., 2008), an extension of AEs, introduce noise into the input data and train the network to reconstruct the original, undamaged signal, thus enhancing the model's robustness to data. Moreover, the concept of autoencoders has evolved into several variants, such as Variational Autoencoders (VAEs) (Kingma, et al., 2013), which model latent variables as probability distributions, providing a new perspective for learning generative models, although they may encounter mode collapse issues during training. Some researchers reviewed the applications and challenges of AEs in machine learning, pointing out that existing methods still face high computational costs when handling high-dimensional data.

## 2.3. ARCHITECTURAL FLOOR PLAN GENERATION

In the field of architecture, the automatic generation of floor plans represents a new research direction aimed at improving design efficiency and quality through computational methods. In related research, some researchers (Hu, et al., 2020) proposed Graph2Plan, a deep neural network that combines generative modeling and user input to automatically generate floor plans that meet specific layout and boundary constraints. some researchers (Zeng et al., 2024a, 2024b) proposed HouseDiffusion, a diffusion-based method for generating vectorized floor plans. While this method improves the precision and diversity of generated plans, it encounters challenges when handling highly complex layouts. Some researchers (Nauata, et al., 2021) developed House-GAN++, a layout refinement generation method based on Generative Adversarial Networks (GANs). While it enhances the diversity and precision of generated layouts , its training process is highly complex without the support of a rich database of design iterations. Compared to existing architectural work (Lu, et al., 2016;

Li et al., 2015, 2019) , our method focuses on the reconstruction and generation of architectural floor plans, enabling more accurate recognition and utilization of the complex abstract information in architectural plans.

## 3. Methodology

In this section, we will provide a detailed introduction to the architectural design, dataset construction, and training process of FloorplanMAE. Our FloorplanMAE model consists primarily of two parts: a lightweight encoder and an asymmetric decoder.

- FloorplanMAE Encoder: Our encoder is based on ViT and operates only on the visible, unmasked image patches. It embeds patches by applying linear projection and adding positional embeddings., and then processes the resulting set through a series of Transformer blocks.
- FloorplanMAE Decoder: The input to the decoder is the complete set of tokens, including both the encoded visible image patches and the mask tokens. Each mask token is a shared, learnable vector, which indicates the target of the prediction task. We add positional embeddings to all tokens in the full set; without doing this, the mask tokens would have no information about their position within the image. The decoder processes this full set with another series of Transformer blocks.

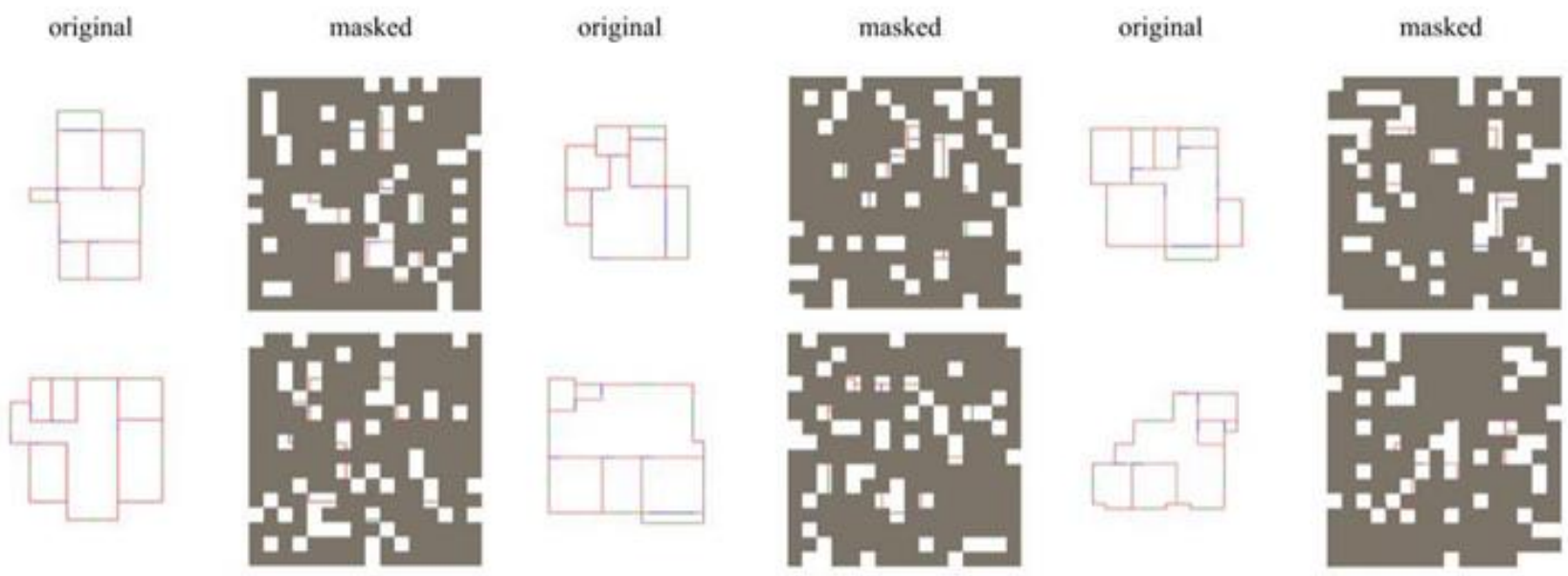

*Figure 2. Random Masking Method.*

The core goal of FloorplanMAE is to restore complete floor plans from partial ones, simplifying the frequent task of modifying floor plans during the architectural design process and intuitively inspiring architects in a fast manner. The FloorplanMAE framework for architectural floor plan reconstruction adopts a straightforward autoencoding method, reconstructing new image signals from partially obtained ones. Our loss function can be defined as follows:

$$L_{\mathrm{MAE}} = \frac{1}{N_{\mathrm{mask}}} \sum_{i \in \mathrm{mask}} (x_i - \hat{x}_i)^2$$

Where $x_i$ represents the i-th masked input data, $\hat{x}_i$ denotes the model's prediction for the i-th data point and $N_{mask}$ indicates the number of masked input points.This loss function measures the mean squared error between the model's predictions and the original masked data. By minimizing this loss function, we can effectively update the model parameters, allowing the model to gradually learn how to reconstruct the masked portions of the images, thereby generating more accurate complete architectural floor plans.

During training, we employ a self-supervised learning approach, training the model to predict the missing parts using only partial floor plan inputs, as Figure 2 and Figure 3 show. The dataset is divided into training, validation, and test sets, containing 7000, 500, and 500 floor plans, respectively.

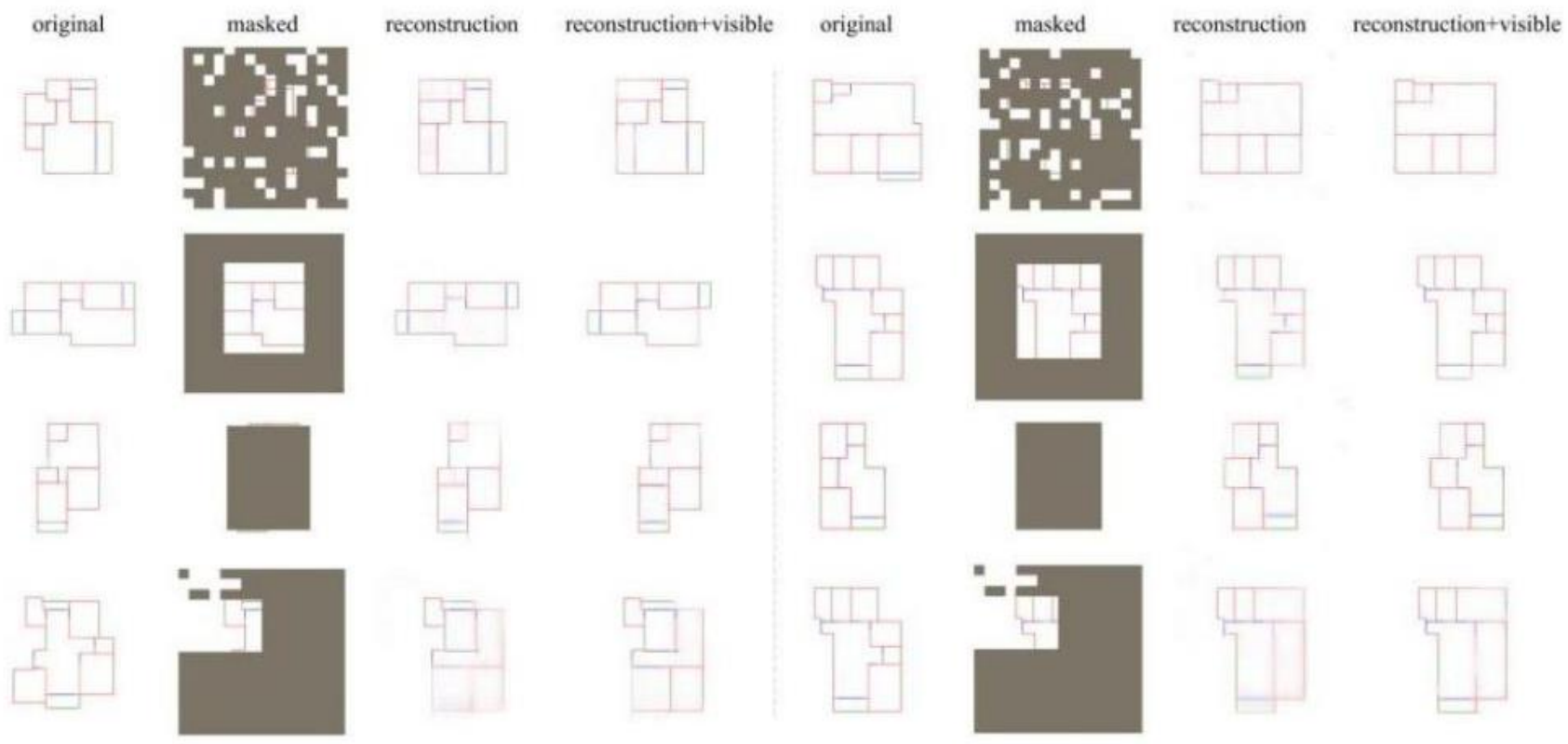

*Figure 3. Example Results_Comparison of Different Masking Strategies.*

We divide architectural floor plans into regular, non-overlapping image patches. We then sample a portion of these patches and mask the remaining ones. Our sampling strategy starts with random sampling: small patches are randomly selected without replacement, following a uniform distribution.The advantages of using random sampling are as follows:

- High random sampling and masking reduce redundancy, turning patch reconstruction into extrapolation from masked neighbors.
- Uniform distribution avoids central bias, but diverse sampling strategies are used due to uneven information distribution in floor plans.

## 4. FloorplanNet Experiments

### 4.1. SELF-SUPERVISED PRETRAINING

The FloorplanNet dataset we utilized consists of 300 collected Real-plan floor plans and 7700 standardized R-plan floor plans, which include colored plans and line drawings. This dataset pertains to residential architectural floor plans, with standardized layouts generated by the RPLAN toolbox. Different colors represent various architectural components, as illustrated in the figure 4.

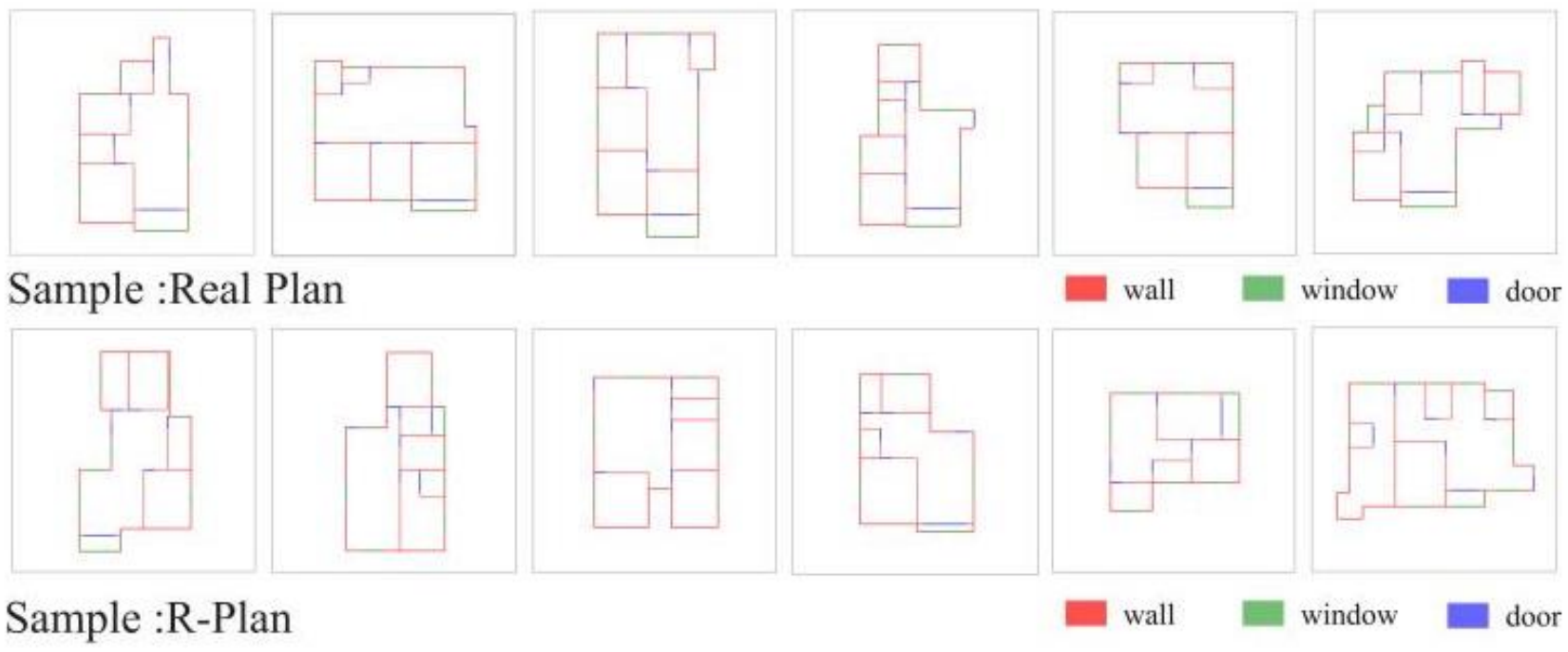

*Figure 4. Illustration of the line drawing floor plan dataset.*

The colored floor plans, as shown in the figure 5, depict different residential layouts, with various colors and shapes denoting different room types. Both datasets contain detailed labels and metadata, ensuring the broad applicability of the test results.

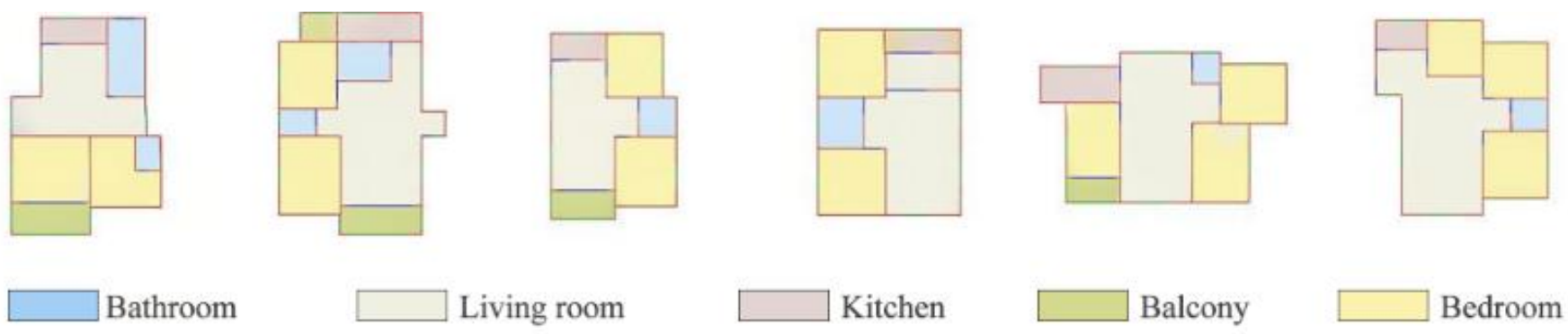

*Figure 5. Illustration of the colored drawing floor plan dataset.*

We performed self-supervised pretraining on the FloorplanNet dataset, which is a critical component of the FloorplanMAE framework. We processed each image by masking part of the floor plan using different masking methods and creating a new training set, with specific sampling strategies detailed in section 4.2.

## 4.2. MASKING AND SAMPLING STRATEGIES

We employed multiple masking strategies, including random, center, edge, and semantic-based sampling. Random sampling masks floor plan patches randomly, forcing the model to infer the structure from limited information. For architectural floor plans, a higher masking ratio (e.g., 75%) improved the model's understanding of the overall structure, unlike natural images, which use lower ratios.

Since architects often sketch continuous sections first, we also tested center, edge, and one-sided sampling to let the model infer missing areas from larger, contiguous patches. Results showed that image quality depended more on patch distribution than content. Greater distances between masked and known patches led to blurrier but easier reconstructions.

## 4.3. QUALITATIVE EXPERIMENTAL RESULTS

In this section, we present the reconstruction results of FloorplanMAE under different masking strategies and compare its performance under varying conditions. First, we experimented with the following masking strategies: random masking, center masking, and edge masking. Below is an analysis of the experimental results for each strategy.

### *4.3.1. Random Masking*

The random masking strategy involves randomly selecting and masking parts of the floor plan to evaluate the model's reconstruction ability under uneven information distribution. In this strategy, the masking ratio of the gray blocks is relatively high, at 80.0%, making it a significant challenge for FloorplanMAE due to the large amount of masked image information. Nevertheless, the model is still able to restore a reasonable architectural floor plan structure and make local adjustments. Although some details, such as the positions of doors and windows and room dimensions, show deviations, the primary functional zoning and overall layout remain clear, as illustrated in the figure 6 and figure 7.

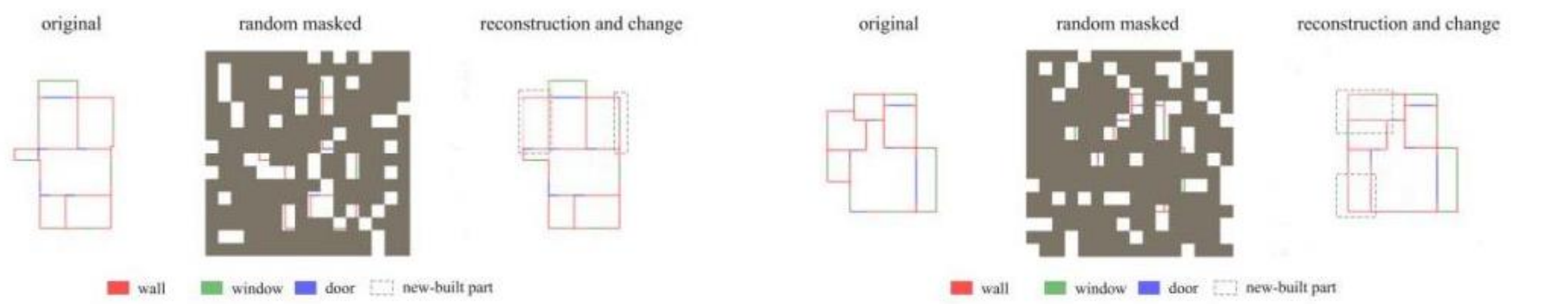

*Figure 6. Detailed Diagram of Random Masking - Line Drawing*

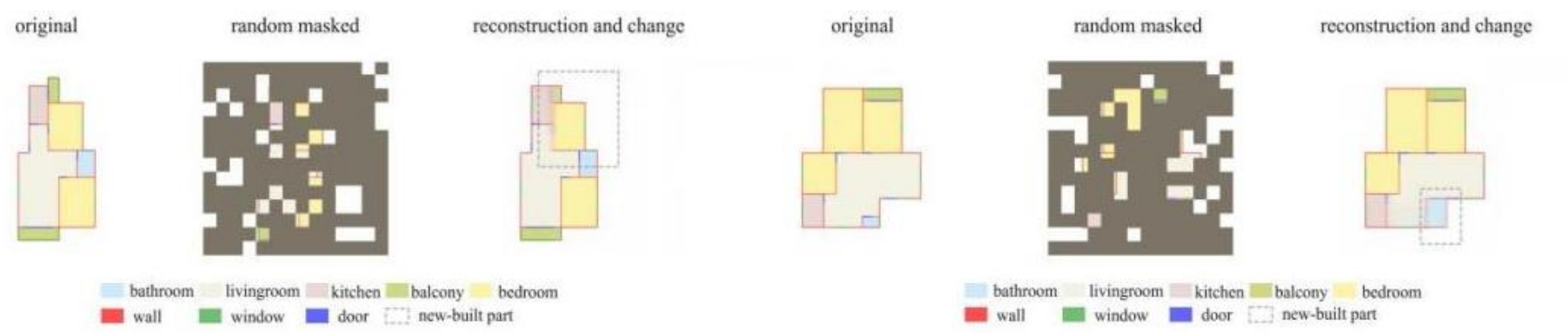

*Figure 7. Detailed Diagram of Random Masking - Colored Floor Plan.*

### *4.3.2. Center Masking*

The center masking strategy masks 30% of the central area of the floor plan to test the model's ability to reconstruct from missing concentrated information. Despite the relatively low masking ratio, key central details are removed, leaving only the floor plan's outline. Results show that FloorplanMAE accurately restores the main layout and structure, though room layouts and door/window positions vary more compared to other strategies. Notably, the model effectively reconstructs crucial architectural elements such as walls and main corridors, as shown in Figure 8.

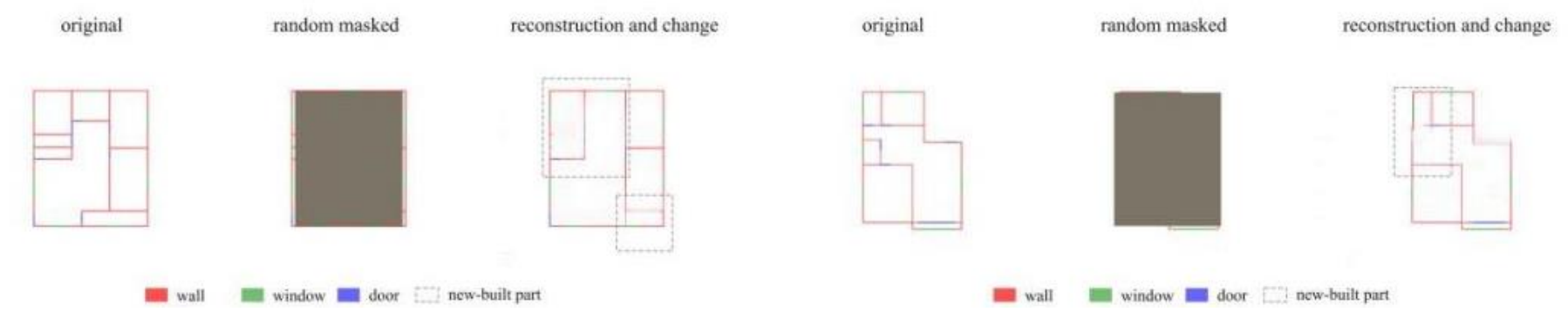

*Figure 8. Detailed Diagram of Center Masking.*

### *4.3.3. Perimeter Masking*

The perimeter masking strategy involves masking the outer parts of the floor plan to simulate scenarios in architectural design where external details may be overlooked. In this strategy, the masking ratio of the gray blocks is 70.0%, covering the entire floor plan. Experimental results show that FloorplanMAE is still able to effectively restore the overall layout and main functional zones of the floor plan under this strategy. Additionally, there is a certain degree of variation in the reconstruction of rooms located at the edges, as illustrated in the figure 9.

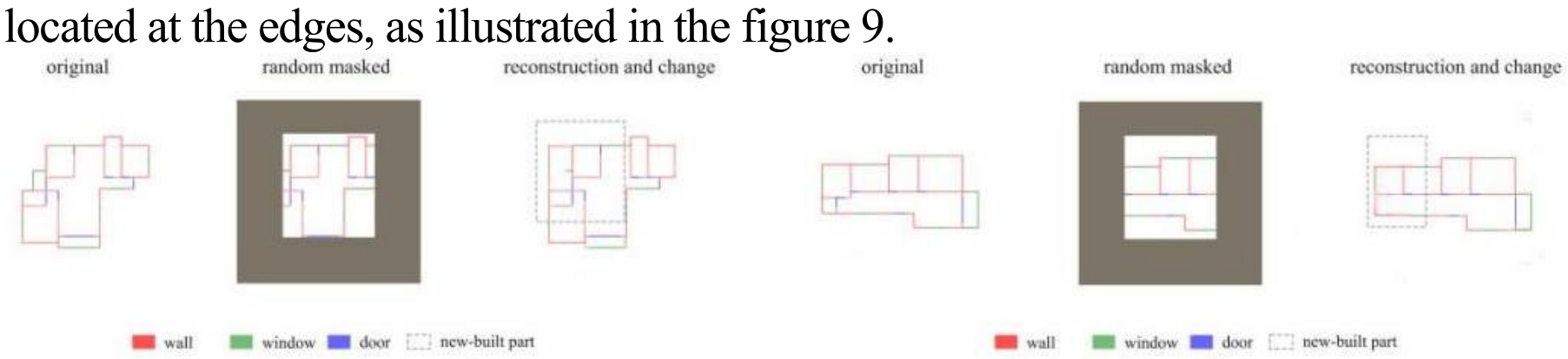

*Figure 9. Detailed Diagram of Perimeter Masking.*

### *4.3.4. One-sided Masking*

The one-sided masking strategy involves masking only one side of the floor plan, simulating scenarios in architectural design where one side of the layout needs to be completed. In this strategy, the masking ratio of the gray blocks is 30%, with minimal coverage of room layout information. As a result, FloorplanMAE can easily restore most of the masked portions of the floor plan. Since most of the original information remains intact, the model performs well in reconstructing both details and the layout,as illustrated in the figure 10.

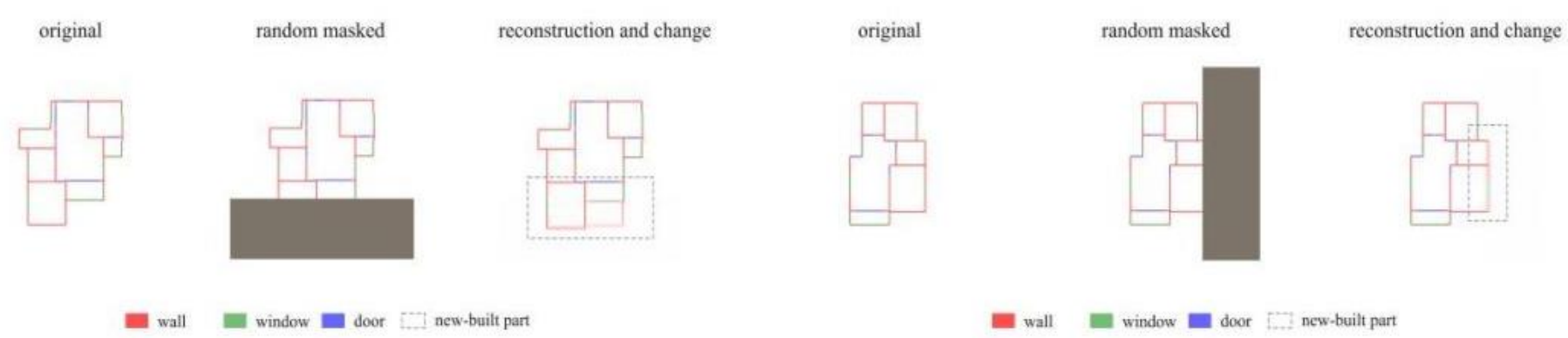

*Figure 10. Detailed Diagram of One-Sided Masking.*

### *4.3.5. Corner Masking*

The corner masking strategy involves masking only one side of the floor plan, simulating scenarios in architectural design where one side of the layout needs to be completed. In this strategy, the masking ratio of the gray blocks is 75%, leaving only a small portion of the rooms visible. Despite this, the model is still able to restore a reasonable architectural floor plan structure, although significant changes occur. The walls and windows in the corner areas exhibit relatively greater blurriness, but the reconstructed rooms still maintain reasonable layout and functional zoning, as illustrated in the figure 11.

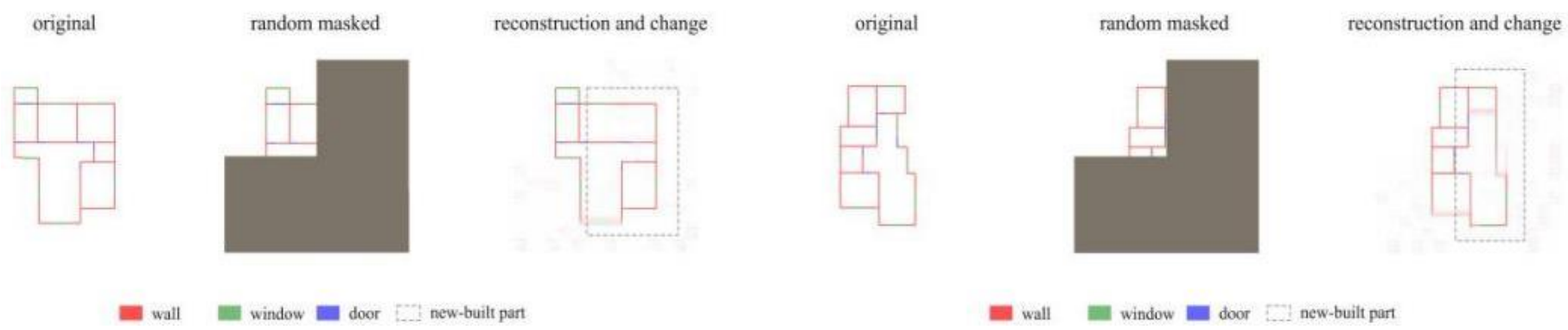

*Figure 11. Detailed Diagram of Corner Masking.*

## 4.4. QUANTITATIVE EXPERIMENTAL RESULTS

We selected pix2pix and cycleGAN as our baselines, and it can be seen from the table 1 that our method outperforms them on various metrics. Overall, both the masking ratio and masking strategy influence the reconstruction outcome. As the masking ratio increases, reconstruction errors tend to rise, and the masking strategy affects the clarity and degree of variation in the reconstructed areas. This further demonstrates the powerful reconstruction capability of FloorplanMAE under conditions of highly incomplete input.

## 5. Discussion and Conclusion

Table 1. Results on FloorPlanNet. The best results are indicated in bold.

| Method | DATA | FID | PSNR | SSIM |
|---|---|---|---|---|
| Center Masking | Line Drawing | 22.0109 | 77.7863 | 0.9738 |
| Corner Masking | Line Drawing | 27.3668 | 76.0285 | 0.9247 |
| One-sided Masking | Line Drawing | **8.84111** | **80.1424** | **0.9794** |
| Perimeter Masking | Line Drawing | *12.4044* | 78.6158 | 0.9609 |
| Random Masking | Line Drawing | 12.6053 | *79.7688* | 0.9757 |
| Random Masking | Colored Drawing | 17.2192 | 79.7638 | 0.9724 |
| Pix2Pix | Line Drawing | 76.3275 | 68.3546 | 0.9334 |
| CycleGan | Line Drawing | 90.3235 | 64.3487 | 0.9468 |

Our results show that FloorplanMAE excels in reconstruction, demonstrating the potential of self-supervised learning for architectural floor plan generation. By predicting missing parts, the model learns the image data distribution, introducing a novel image completion framework and a scalable approach for floor plan processing. Unlike traditional methods where architects draw line by line, FloorplanMAE generates complete designs from partial plans, saving significant time. We also raise an open question about balancing controllability and creativity: do common metrics like FID accurately reflect the rationality of generated designs? We aim to explore this in future work.

FLOORPLANMAE - A SELF-SUPERVISED FRAMEWORK FOR COMPLETE FLOORPLAN GENERATION FROM PARTIAL INPUTS

FLOORPLANMAE - A SELF-SUPERVISED FRAMEWORK FOR COMPLETE FLOORPLAN GENERATION FROM PARTIAL INPUTS